\documentclass{article}

\usepackage[utf8]{inputenc} % allow utf-8 input
\usepackage[T1]{fontenc}    % use 8-bit T1 fonts
\usepackage{hyperref}       % hyperlinks
\usepackage{url}            % simple URL typesetting
\usepackage{booktabs}       % professional-quality tables
\usepackage{amsfonts}       % blackboard math symbols
\usepackage{nicefrac}       % compact symbols for 1/2, etc.
\usepackage{microtype}      % microtypography
\usepackage{xcolor}         % colors
\usepackage{times}
\usepackage{epsfig}
\usepackage{graphicx}
\usepackage{amsmath, amsthm}
\usepackage{amssymb}
\usepackage{multirow}
\usepackage{bbm}
\usepackage[numbers]{natbib}
\usepackage[preprint]{neurips_2023}
\usepackage{subcaption}
\newtheorem{theorem}{Theorem}

\title{Structured Network Pruning by Measuring Filter-wise Interactions}

\author{%
   Wenting Tang \\
   Beijing Key Laboratory of Digital Media, \\
   School of Computer Science and Engineering, \\
   Beihang University\\
   Beijing, China \\
   \texttt{wtang13@buaa.edu.cn} \\
   \And
   Xingxing Wei \\
   Beijing Key Laboratory of Digital Media, \\
   School of Computer Science and Engineering, \\
   Beihang University\\
   Beijing, China \\
   \texttt{xxwei@buaa.edu.cn} \\
   \And
   Bo Li\\
   Beijing Key Laboratory of Digital Media, \\
   School of Computer Science and Engineering, \\
   Beihang University\\
   Beijing, China \\
   \texttt{boli@buaa.edu.cn} \\
}

\begin{document}

\maketitle

\begin{abstract}
Structured network pruning is a practical approach to reduce computation cost directly while retaining the CNNs' generalization performance in real applications. However, identifying redundant filters is a core problem in structured network pruning, and current redundancy criteria only focus on individual filters' attributes. When pruning sparsity increases, these redundancy criteria are not effective or efficient enough. Since the filter-wise interaction also contributes to the CNN's prediction accuracy, we integrate the filter-wise interaction into the redundancy criterion. In our criterion, we introduce the filter importance and \emph{filter utilization strength} to reflect the decision ability of individual and multiple filters. Utilizing this new redundancy criterion, we propose a structured network pruning approach SNPFI (\textbf{S}tructured \textbf{N}etwork \textbf{P}runing by measuring \textbf{F}ilter-wise \textbf{I}nteraction). During the pruning, the SNPFI can automatically assign the proper sparsity based on the \emph{filter utilization strength} and eliminate the useless filters by filter importance. After the pruning, the SNPFI can recover pruned model's performance effectively without iterative training by minimizing the \emph{interaction difference}. We empirically demonstrate the effectiveness of the SNPFI with several commonly used CNN models, including AlexNet, MobileNetv1, and ResNet-50, on various image classification datasets, including MNIST, CIFAR-10, and ImageNet. For all experimental CNN models, nearly 60\% of computation is reduced in a network compression while the classification accuracy remains.

%To maintain the performance of the pruned model, we quantify the  generalization gap between the pruned model and the original model by the \emph{interaction difference}

\end{abstract}

\section{Introduction}
Network pruning can transfer the pre-trained heavy CNNs models to a lightweight form with comparable precision by removing models' inherent redundancy \cite{asif2019ensemble}. On the one hand, this transformation effectively boosts the real-time performance of CNNs on the edge device. On the other hand, network pruning prevents over-fitting by directly eliminating unnecessary parameters \cite{Heaton2017Ian}. Structured network pruning removes a group of parameters in various granularity (e.g.kernel, filter). Since network pruning at the filter level of granularity requires no special-designed hardware accelerator, filter-wise structured network pruning becomes one of the research focuses currently \cite{anwar2017structured, he2019filter,zmora2019neural}.  

\begin{figure}[ht]
    \centering
    \includegraphics[width=1.0\textwidth, height=0.368\textwidth]{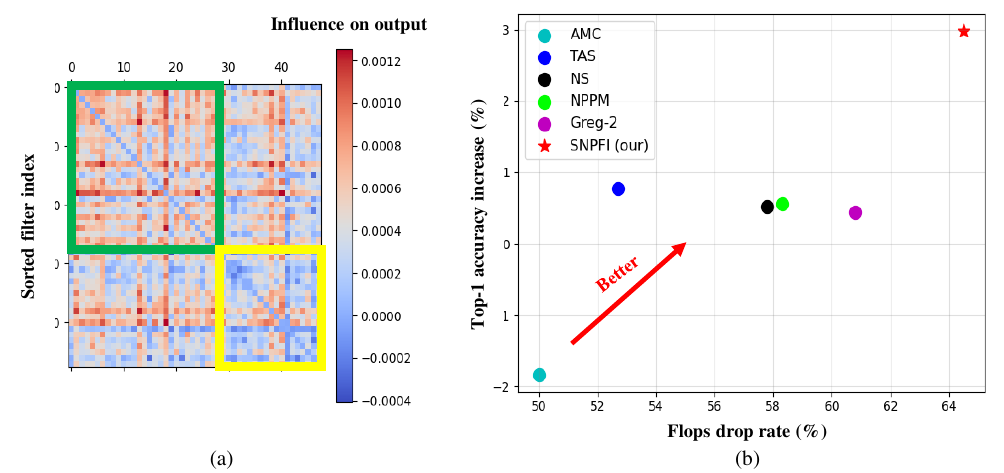}
    \caption{(a) Parts of binary filter interaction of the $123$-th layer of the Resnet-50 \cite{he2016deep} when inference on CIFAR-10 \cite{krizhevsky2009learning}. The filters are sorted by the L1-norm of weight in descending order\cite{han2015learning}. The binary interactions among the important filters are in the green square and the interactions among the subtle filters are in the yellow square; filters' influences on output are computed by \cite{singh2018hierarchical} and normalized.  (b) The compression performance on CIFAR-10 by state-of-the-are pruning algorithms \cite{gao2021network,liu2017learning,2019arXiv190509717D,he2018amc,2021Neural} and SNPFI (our). The red arrow indicates better compression results.} % The SNPFI outperforms the other pruning algorithms by considering the filter-wise interaction during the redundancy evaluation. }
    \label{fig:cifar10-pruning}
\end{figure}

Identifying redundant filters is a core problem in structured network pruning. The filter is redundant if its contribution to the performance is subtle. Because either the filter’s weight or output feature can be the clue to discriminate its redundancy, structured network pruning algorithms can be categorized as weight-based \cite{lecun1989optimal,Zhong2022RevisitKP,han2015learning} and feature-based approaches \cite{Zou2018ConvolutionalNN, liu2017learning, bu2021learning}. Assuming the vital filters have large weights, the LWCE \cite{han2015learning}  identifies redundant filters by the L1-norm. Considering the similarity among filters in feature, the FPGM \cite{he2019filter} ranks filters by the difference from the geometric median. When pruning intensity is modest, these static criteria can efficiently identify redundancy within a single evaluation. Thus, they are also known as \emph{one-shot pruning}. Under high pruning intensity, one-shot pruning approaches struggle to recover the performance effectively. Dynamic evaluation overcomes this drawback by measuring redundancy with more caution. \citet{DBLP:conf/iclr/YeL0W18} searches the redundant filters multiple times and recovers the weight through iterative training. Current feature-based approaches dynamically evaluate filters during the inference and the training \cite{liu2017learning, gao2021network}. The Network slimming \cite{liu2017learning} embeds the filter selector layers into the original CNN to improve the accuracy of the redundancy discrimination. Despite various redundancy criteria existing, how to effectively and efficiently identify redundancy is still an unsettled problem.

Solely focusing on filter’s unitary attributes, the above redundancy criteria neglect the filter-wise interaction when measuring filter's contribution on prediction. Many network interpretation studies have shown that filter-wise interactions do contribute on the prediction \cite{DBLP:conf/iclr/TsangC018,janizek2021explaining,Cui2019LearningGP}. On the one hand,  each filter's interaction can reflect its inherent decision ability \cite{janizek2021explaining}. On the other hand, the filter-wise interactions are universal \cite{singh2018hierarchical} and the functionally close-coupled layer exists in common CNNs as shown in Figure \ref{fig:cifar10-pruning}-(a). In this scenario, filters with various importance have to interact to positively contribute to the output. Therefore, these motivate us to integrate the filter-wise interaction into our redundancy criterion . As shown in Figure \ref{fig:cifar10-pruning}-(a), important filters (green cube) can benefit considerably from interacting with the subtle filters (yellow cube). Thus, only measuring the unitary attribution of filters can not effectively identify redundancy. Since the useless filters are functionally loosely-coupled (blue region) in the layer, measuring the strength of interaction among multiple filters can avoid excluding the filters that seem to be useless but contribute.  
%In our redundancy criterion, the filter is redundant only if it contributes little to the peer filters and the output. 

To identify the functionally loosely-coupled filters, we measure the filter's importance and the utilization strength of interaction by the filter-wise interaction in our redundancy criterion. The contributions of individuals or groups of filters on output reflect the importance of each filter and the strength of interaction.  To distribute the unit and the group contribution fairly, we model the filter-wise interaction by the \emph{cooperative game theory} \cite{chalkiadakis2012cooperative}. For the different number of filters in each layer, we refer to the normalized strength of interaction among multiple filters as the \emph{filter utilization strength}. Since the filter utilization strength can reflect the decision ability of a group of filters, we can provide a theoretical lower bound of sparsity concerning the performance. What's more, we identify the interactions that might cause the generalization gap by comparing the interaction result of the pruned model and the original model. For each pruned computational module, we refer to this disparity as the \emph{interaction difference} to quantify the potential generalization gap and fine-tune the pruned model with the interaction differences to boost the optimization effectively. In this way, we can maintain the performance of the pruned model by our redundancy criterion.
%is naturally related to the number of filters and their cooperative contribution to the inference
% \begin{figure}[h]
% \begin{minipage}[t]{0.4\textwidth}
%     \centering
%     \includegraphics[width=1.0\textwidth]{cifar10-pruning.pdf}
% \end{minipage}
% \begin{minipage}[t]{0.65\textwidth}
%     \centering
%     \includegraphics[width=0.8\textwidth]{NPFI.pdf}
% \end{minipage}
% \caption{left: NPFI (square) and SOTA pruning algorithms' compression result on CIFAR-10 for ResNet-50 and MobileNetv1. right: The architecture of the NPFI. }
% \label{fig:NPFI}
% \end{figure}

\begin{figure}[ht]
    \centering
    \includegraphics[width=1.0\textwidth]{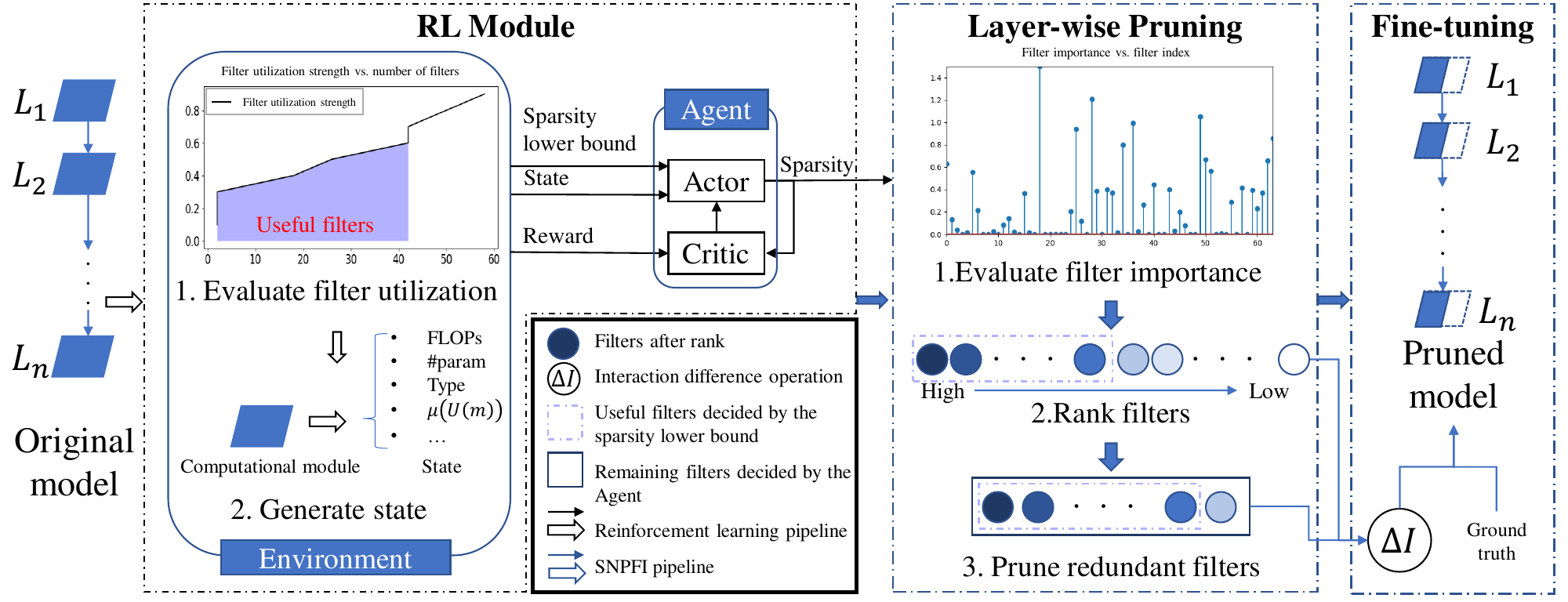}
    \caption{Overview of the proposed SNPFI. The blue parallelogram is the computational module before and after pruning. Each computational module consists of a few filters (blue circle). According to our redundancy criterion, the RL module can predict the proper layer-wise sparsity efficiently by the filter utilization strength; the Layer-wise pruning module can identify the relative redundant filters accurately by the filter importance; the Fine-tuning module can recover performance effectively by the interaction difference.}
    \label{fig:NPFI}
\end{figure}

To attain the optimal combination of the layer-wise pruning sparsity, we propose a structured network pruning approach SNPFI (Structured network pruning by measuring filter-wise interaction) in Figure \ref{fig:NPFI} based on our redundancy criterion. SNPFI includes three parts: the RL (reinforcement learning) module, the Layer-wise Pruning module, and the Fine-tuning module. The RL module predicts the pruning sparsity in layer-wise and consists of the Environment and the Agent. For each computational module, the Environment first generates the state and the sparsity lower bound via the filter utilization strength; then the Agent predicts the sparsity by the state and the sparsity lower bound; in the end, the Environment rewards this pruning decision in constant time. Once the pruning is finished, the Agent is updated by the pruning decision and reward. To alleviate the \emph{delayed reward} problem of model-free reinforcement learning in the pruning scenario \cite{arjona2019rudder}, we formulate the reward function by filter utilization strength. In this way, SNPFI can automatically and efficiently optimize the compression plan; The Layer-wise pruning module removes the redundant filters based on the predicted sparsity. During the layer-wise pruning, the redundant filters are removed via the filter importance and the sparsity; The Fine-tuning module recovers the weight of the pruned model according to the interaction difference. In this way, we can ensure the pruned model acquires meaningful interaction behaviors inherent in the original model even under a high pruning intensity. As shown in Figure \ref{fig:cifar10-pruning}-(b), SNPFI can outperform the other state-of-the-art pruning approach even with a more limited computation overhead.

The contributions of our work are as follows:
\begin{itemize}
    \item We propose a redundancy criterion by filter-wise interaction and theoretically and experimentally prove its effectiveness. In this new redundancy criterion, the filter importance and filter utilization strength imply the decision ability of individual and multiple filters. According to our redundancy criterion, the interaction difference abstracts the potential generalization gap due to the pruning and effectively guides the weight recovery of the pruned model;
    \item We present a model compression algorithm SNPFI which prunes efficiently and effectively; Utilizing our redundancy criterion and the interaction difference, SNPFI avoids pruning the relatively useful filters and recovers the performance by minimizing the generalization gap caused by pruning;
    \item We experimentally demonstrate the availability of SNPFI among MNIST, CIFAR10, and IMAGENET across AlexNet, MobileNetv1, and ResNet-50. SNPFI can always prune around 60\% (20\% higher than experience) of filters of the original model without significant performance decline at a single-pass pruning process.
\end{itemize}

%------------------------------------------------------------------------
\section{Related work}

\textbf{Weight-based and Feature-based network pruning.} Network pruning is a process that decouples the useless structure from the CNN by the redundancy criteria. The unitary attributes (e.g.weight, feature) can spot redundancy well under a conservative pruning sparsity \cite{lecun1989optimal, he2019filter, Zhong2022RevisitKP}. The OBD \cite{lecun1989optimal} measures neurons' importance by the second-order derivative of weight; The TMI-GKP \cite{Zhong2022RevisitKP} evaluates the redundancy by the similarities among filters by weight. Since the coupling degrees of different computational modules might vary in diverse applications \cite{singh2018hierarchical, janizek2021explaining}, static evaluations of redundancy are not applicable when pruning intensity increases \cite{DBLP:conf/iclr/YeL0W18}. Therefore, current feature-based pruning approaches design filter selectors to decouple the useless filters at the expense of the computation\cite{ gao2021network,liu2017learning}. The NPPM \cite{gao2021network} can foresee the future impact on accuracy for any layer-wise pruning, but requires extra training on the filter selector. Thus, how identifying redundancy effectively and efficiently under the high pruning intensity is still an unsettled problem. We address this problem with a new redundancy criterion by the filter-wise interaction in Section \ref{subsec:3.1}.

\noindent\textbf{Structured pruning and AutoML.} 
Structured pruning \cite{anwar2017structured} removes groups of neuron connections at once, which means the adjacent and preceding layer's output channels shrink with the current layer's input channels when pruning at filter granularity. Since the computation cost exponentially decreases when scheduling the model's compression plan at the filter level, AutoML \cite{yao2018taking} (Automated machine learning) can effectively assign layers' sparsity by repeated retraining on each pruned model  \cite{lin2020channel}. Since repeated retraining is necessary to bond the accuracy and the sparsity, the reinforcement learning approaches also struggle to balance exploration and exploitation due to the delayed reward problem \cite{he2018amc} or lack of exploration \cite{yu2022topology}. The delayed reward problem is the reinforcement learning tasks with sparse or episodic rewards \cite{arjona2019rudder}. Aiming to alleviate the delayed reward problem in real pruning scenarios, we formulate the reward function by filter utilization rate in Section \ref{subsec:3.3}.

\section{Method}

To identify the redundancy effectively, we introduce a new redundancy criterion by the filter-wise interaction in Section \ref{subsec:3.1}; To demonstrate the feasibility of our redundancy criteria, we propose a reinforcement learning-based structured network pruning approach SNPFI in Section \ref{subsec:3.3}; To minimize the performance disparity between the original and the pruned model, we propose the interaction difference for fine-tuning in Section \ref{subsec:3.4}. %Considering the interactions among image bands, we analyze the model redundancy in different applications in Section \ref{subsec:3.2}.

\subsection{Measuring the redundancy}\label{subsec:3.1}

In the pruning scenario, there are two types of filters in each computational module: the redundant and the useful. Taking a CNN with $n$ computational layers as an example, the $l$-th layer is a set of filters $N_l=\left\{F_i^l|i=1,...,c_{out}^l\right\}$, where each filter is $F_i^l\in \mathbb{R}^{C_{in}^l\times k^l\times k^l}$, $c_{out}^l$ and $c_{in}^l$ are the $l$-layer's output and input channels, $k^l$ is the kernel size. The layer's pruning sparsity indicates the ratio of number of the redundant filters out of all the filters $N_l$ \cite{anwar2017structured}. According to the redundancy criteria and the $l$-th layer's pruning sparsity, assume $r^l$ filters are useful, then the pruning sparsity is $\frac{c_{out}^l-r^l}{c_{out}^l}$. The useful filter set $S_l$ of the $l$-th layer is $S_l=\left\{F_i^l|i \in IMP \right\}$, where the $IMP$ is a set of the index of the useful filters identified by their importance according to the specific redundancy criteria. In the inference, each filter consumes a comparable amount of computation but contributes to the output differently. Since the redundant filters' contributions are much lower than the useful filters, current pruning algorithms believe that the influence of pruning the redundant filter is subtle\cite{he2019filter, 2016DeepCompression, anwar2017structured}. However, these approaches neglect the filter-wise interactions' contributions during the redundancy evaluation. 

Since a relatively useless filter might have a considerable collaborative contribution to prediction \cite{janizek2021explaining, singh2018hierarchical}, we need to fairly quantify the impacts of the potential filter-wise interactions on prediction. To achieve this goal, we regard each inference process achieved by $m$ filters in the $l$-th layer as a \emph{collaborative game} $<M_l, V>$ \cite{Grabisch1999AnAA}. During the inference on a image $I$, each filter is a \emph{player} and $m$ players align the \emph{coalition} $M_l$ with the contribution $V(M_l)$, where $m = |M_l| \in [1,c_{out}^l]$, $V(M_l)=log \frac{P(\hat{y}=cls|M_l,I)}{1-P(\hat{y} =cls |M_l,I)}$ \cite{deng2021discovering} and $\hat{y}$ is the prediction with $M_l$. According to the $V(M_l)$ for each coalition $M_l$, we can distribute the contributions on the output when multiple filters interact. In this way, we can quantify each filter's importance and each layer's \emph{filter utilization strength} to identify the functionally loosely-coupled filters. To fairly measure the individual contributions of each filter, we formulate the importance of the $c$-th filter in the $l$-th layer by the Shapley value \cite{rozemberczki2022shapley} in Eq.(\ref{eq:triviality}).
\begin{equation} \label{eq:triviality}
    t_c^l = \left|\sum_{c\in M_l,M_l\subseteq N_l }\frac{|M_l|!(c_{out}^l + 1 -|M_l|)!}{c_{out}^l!} \bigtriangleup V(M_l,c) \right|,
\end{equation}
where $\bigtriangleup V(M_l,c)=V(M_l) -V(M_l -\left\{ c\right\})$. When the $c$-th filter forms the $M_l$ with other filters in the $l$-th layer, it might positively contribute to the prediction. Considering the potential contribution of each filter, the filter importance $t_c^l$ assigns the relatively useless filter to a small value. As proven in \cite{ancona2020shapley}, the $t_c^l$ is unique and fair to each filter. Therefore, we can fairly discriminate the relative useless filter from the useful filter.

To fairly distribute the contribution of interaction, we first introduce the filter interaction $u_l^{d}(i,j)$.
The filter interaction is the contribution of the interaction among at least two distinct filters. In this scenario, coalition $M_l$ has to consist of $m=d+2$ filters, where $\left\{ i, j\right\}\subseteq M_l$, $i,j \in [1, c_{out}^l]$, $i\neq j$, $d \in [0, c_{out}^l-2]$ and $m \in [2, c_{out}^l]$. Utilizing the Shapley interaction index \cite{Grabisch1999AnAA}, the filter interaction $u_l^{d}(i,j)$ among $i,j$, when the other $d$ filters exist, define in Eq.(\ref{eq:ci}).
\begin{equation}\label{eq:ci}
    u^d_l(i,j)=\sum_{\left\{i,j\right\}\subseteq M_l\subseteq N_l, \left | M_l \right|=d+2}\bigtriangleup V(i,j,M_l),
\end{equation}
where $\bigtriangleup V(i,j,M_l)=(V(M_l)-V(M_l-\left \{ j \right \})) - (V(M_l-\left \{ i \right \})-V(M_l-\left \{ i,j \right \}))$. The larger  $u^d_l(i,j)$, the stronger interaction when $i$,$j$ form a coalition with the other $d$ filters. Notably, it can be proved that $u^d_l(i,j)$ is unique and fair among all coalitions\cite{rozemberczki2022shapley}. With the $u^d_l(i,j)$, we can measure the filter utilization strength $U_l(m)$ of the $l$-th layer in Eq.(\ref{eq:cu}).
\begin{equation}\label{eq:cu}
    U_l(m)=\frac{\sum_{q=0}^{m-2}\mathbb{E}_{I\in\Omega }\mathbb{E}_{i,j\in N_l}[u^{q}_l(i,j)]}{\sum_{p=0}^{c_{out}^l - 2}\mathbb{E}_{I\in\Omega }\mathbb{E}_{i,j\in N_l}[u^{p}_l(i,j)]},
\end{equation}
where the $\Omega$ is the calibration dataset and $|\Omega|=256$ by default. A high value of $U_l(m)$ indicates that the interaction strength is intensive when $m$ filters exist. If subtle filters exist, $U_l(m)$ achieves a high value with a relatively small $m$. In this way, we can estimate the number of useless filters by $U_l(m)$.

% \begin{itemize}
%by satisfying the linearity, efficiency, symmetry and null interaction
%     \item \textbf{Linearity.} If the output of the network can be seen as a linear combination of the output of two sub-network, then any attribution should also be a linear combination, with the same weights, of the attributions computed on the same sub-networks. Because linearity has been justified as an axiom of the deep neural network in \cite{2017Axiomatic}, $u^m(i,j)$ satisfies linearity by using the output of CNN to formulate characteristic function.
%     \item \textbf{Efficiency.} The $V'$ sum up to the difference between $V'(\emptyset)$ and $V(N\setminus \left \{ i,j \right \})$ where $ \left \{i\right \}, \left \{j\right \}$ can be $\emptyset$. 
%     \item \textbf{Symmetry.} If $i,j,x,w\in N$ that $V( S\cup\left \{ i \right \})=V( S\cup\left \{ x \right \}) \wedge  V(S\cup\left \{ j \right \})=V(S\cup\left \{ w \right \})\wedge  V(S\cup\left \{ i,j \right \})=V(S\cup\left \{ x,w \right \})$ for all $S\in N\setminus \left \{ i,j,x,w \right \}$, then $u^m(i,j)=u^m(x,w)$.
%     \item \textbf{Null interaction.} If a channel contributes nothing to the output, then the channel interaction with the channel should be zero.
% \end{itemize}

\subsection{Filter-wise interaction based structured network pruning}\label{subsec:3.3}
In the pruning scenario, the approximation of the optimal pruning plan $S^*$ can be time-consuming. Since each layer's pruning sparsity and the accuracy of the pruned model are non-linearly related, previous studies spend considerable amount of computation to sample the best accuracy achieved by the according pruning sparsity\cite{gao2021network,lin2020channel, yu2022topology}.
%here need to demonstrate why our criterion's lower bound is more efficient
Based on our redundancy criterion, we can estimate each layer's sparsity lower bound by the expert experience. Given the desirable filter utilization strength $\theta$ to maintain the basic functionality after pruning, the lower bound of the $l$-th layer's sparsity $s_{lb}^l$ is estimated in Eq.(\ref{eq:slb}).
\begin{equation}\label{eq:slb}
    \min _{m} s_{lb}^l =\frac{m+2}{c_{out}^l}\, s.t. \, U_l(m)\geqslant \theta,\, m \in [2, c_{out}^l ] \land m \in \mathbb{Z}^{+}
\end{equation}
According to the $s_{lb}^l$ and $t_c^l$, the relatively useful and important filters always remain in the layer-wise pruning as shown in the dotted box of Figure.\ref{fig:NPFI}. However, scheduling the layer-wise pruning plan only by the expert experience might compromise with the local optimum \cite{he2018amc, yu2022topology}. The objective function of the scheduling process is in the E.q.(\ref{eq:opt_prune}).
\begin{equation}\label{eq:opt_prune}
    S^*=\min _{S} COMP(S) \, s.t.\, ACC(S) \geqslant \alpha,\,S=\left\{s^l|l=1,...,n\right\},\,s^l \in(0,1.0]
\end{equation}
where $COMP(S)$ and $ACC(S)$ are the computation and the accuracy of the pruned model following the pruning plan $S$, and $\alpha$ is the legal validation accuracy for the pruned model. To efficiently approximate the $S^*$ through enormous combination of $S$, we integrate $s_{lb}^l$ into E.q.(\ref{eq:opt_prune}):
\begin{equation}\label{eq:opt_prune_now}
    S^*=\min _{S} COMP(S) \, s.t.\, ACC(S) \geqslant \alpha,\,S=\left\{s^l|l=1,...,n\right\},\,s^l \in[s_{lb}^l,1.0]
\end{equation}
%The purpose of network pruning is to attain a lightweight but effective network.  Specifically, the pruning plan should decline the computation overhead while maintaining the generalization performance of the pruned model. Thus, we utilize the RL algorithm to optimize the pruning plan based on each layer's sparsity lower bound. When scheduling the pruning sparsity at the filter granularity, the searching space grows exponentially with the number of filters at each layer. This enormous search space might lead to insufficient exploration during the optimization of the RL agent \cite{yu2022topology}. The AMC uses the continuous action space and sets the reward of the intermediate layers to zero to boost the exploration \cite{he2018amc}. However, the sparse rewards can not promise the scheduling process heading to the optimal solution \cite{ruder2016overview}. 
According to E.q.(\ref{eq:opt_prune}) and E.q.(\ref{eq:opt_prune_now}), it is straightforward that the searching space sharply shrinks and is feasible for the RL algorithm. Therefore, we integrate our filter-wise interaction-based redundancy criterion into the RL algorithm as shown in Figure \ref{fig:NPFI}.
In the following, we explain the details of the RL module of the SNPFI.

%As shown in Figure \ref{fig:NPFI}, the RL module predicts the pruning sparsity in layer-wise order. In the RL module, the Agent predicts the pruning sparsity by the lower bound of the layer's sparsity. Formulating the reward function by filter utilization, the Environment can reward the prediction in constant time. In this way, the SNPFI sharply declines the search space and alleviates the delayed reward problem. In the following, we explain the details of the RL module in SNPFI.

%\textbf{Environment} generate state and reward the Agent. For each accessible module (fully connected layer, the convolutional layer, or the building blocks for modern CNNs), the environment first generates the state by filter utilization strength; then suggests a lower bound of the layer's sparsity to the SNPFI Agent; at last, the environment rewards the Agent's action by filter utilization rate in constant time.

\textbf{State $s^l$} is defined below for the $l$-th accessible layer:
\begin{equation}\label{eq:s}
    s^l=<type,\#param, FLOPs,k^l,\mu(U_l(m)), \sigma(U_l(m)), c_{in}^l, c_{out}^l >
\end{equation}
The $type$ is the type of the layer, $\#param$ is the number of the parameters, and $FLOPs$ is the number of floating-point operations. The average and the standard deviation of the filter utilization strength, $\mu(U_l(m))$ and $\sigma(U_l(m))$, are included in the state.

\textbf{Action} $a^l$ is the pruning sparsity of the $l$-th layer. In the Agent, the $a^l$ is predicted by the policy network $\pi_{\epsilon}$ (the Actor in Figure \ref{fig:NPFI}) on the state $s^l$ and bounded by the $s_{lb}^l$. $a^l$ is defined in the Eq.(\ref{eq:action}).
\begin{equation}\label{eq:action}
    a^l=\max(s_{lb}^l,\pi_{\epsilon}(s^l))
\end{equation}
where the $\epsilon$ is the trainable parameters of $\pi$. The output range of the policy network is in $(0, 1]$.

\textbf{Reward} should convey the future effects on the model’s computational overhead and performance caused by the following layer-wise pruning. Since the $U_l(m)$ is naturally related to the number of filters and their cooperative contribution to the inference, we can formulate the reward function $R_l(\cdot)$ of the $l$-th layer via $U_l(m)$ and $r^l$ to alleviate the delayed reward problem \cite{ruder2016overview} in the AMC \cite{he2018amc}:
\begin{equation}\label{eq:reward}
    R_l(r^l)=\left\{\begin{matrix}
\frac{U_l(r^l)}{r^l}-\frac{1}{c_{out}^l} & ,1\leqslant l< n\\ 
 \frac{\sum_{i=1}^{n-1}R_i(r^i)}{n}+ \frac{c_{out}^n\times U_n(r^n)-r^n}{r^n \times c_{out}^n \times n}& ,l=n\wedge ACC(S)\geqslant \alpha \\ 
 \frac{\sum_{i=1}^{n-1}R_i(r^i)}{n} + \frac{c_{out}^n\times U_n(r^n)-r^n}{r^n \times c_{out}^n \times n} -n^{2}& ,l=n\wedge ACC(S)< \alpha
\end{matrix}\right.
\end{equation}
where $r^l=\lfloor c_{out}^l \times a^l \rfloor$ is the number of remaining filters and $S=\left\{a^l|l=1,...,n  \right\}$ is the predicted pruning plan. For CIFAR-10 and MNIST, we evaluate the $ACC(S)$ on the validation set; for ImageNet, we use the training set. When the $S$ ensures the performance of the pruned model, this reward function encourages the $S$ to achieve a higher filter utilization strength with fewer filters during the optimization; When the $S$ violates the legal performance of the pruned model $\alpha$, this reward penalizes the actor for the aggressive pruning sparsity. In this way, the $S\to S^*$ step by step.

\textbf{Policy Update.} We utilize the DDPG \cite{lillicrap2015continuous} algorithm to optimize the pruning policy. The parameters of the policy network $\pi$ are updated by the Eq.(\ref{eq:DDPG}).
\begin{equation}\label{eq:DDPG}
    J(\epsilon)= \mathbb{E}_{s^l\sim \rho ^{\beta }}[Q^{\pi}(s^l,\pi_{\epsilon}(s^l))]
\end{equation}
In DDPG, $\rho ^{\beta }$ is the critic network (the Critic in Figure.\ref{fig:NPFI}) to encourage the right action, $Q^{\pi}$ is the value function to estimate current policy.
\subsection{Interaction difference based fine-tuning}\label{subsec:3.4}
During the deployment on the edge device, removing relatively subtle filters with considerable collaborative contribution might be inevitable. The generalization gap between the pruned and the original model emerges in this scenario. Since our redundancy criterion prevents pruning the filters both useful and important, this gap is caused by the distinct interactions of the pruned filters \cite{janizek2021explaining,Cui2019LearningGP}.  Therefore, we propose to utilize the filter-wise interaction for fine-tuning as shown in Figure.\ref{fig:NPFI}. In the perspective of the collaborative game, the remaining $r^l$ filters and the entire $c_{out}^l$ filters form two different coalitions after pruning on the $l$-th layer: $S_l$ and $N_l$. If the generalization gap exists,  at least one cooperative filter exists in the $\left\{N_l - S_l\right\}$. To quantify this generalization gap, we define the interaction difference $\Delta I(S_l, N_l)$ among $S_l$ and $N_l$ in the Eq.(\ref{eq:ID}).
\begin{equation}\label{eq:ID}
    \Delta I(S_l, N_l) = \mathbb{E}_{(S_l,N_l)}[\frac{r^l}{c_{out}^l} \times V(N_l) -  V(S_l)]
\end{equation}

\begin{theorem}\label{th:ID}
If $V(S_l),V(N_l)\in\mathbb{R}$ is such that $V(S_l)\neq V(N_l)$ for any computational layer indexed by $l$, then $\Delta I(S_l, N_l)>0$. 
\end{theorem}
Based on the theorem \ref{th:ID}, the non-negative interaction indicates the existence of meaningful interactions that leads to a better generalization. Hence, we propose the ID loss in Eq.(\ref{eq:ID loss}) to encourage the pruned model to learn these important interactions.
\begin{equation}\label{eq:ID loss}
    L_{ID}=-\frac{1}{n} \sum_{l=1}^{n}\sum_{cls=1}^{\mathbb C}P(\hat{y}=cls|\Delta I(S_{l}, N_l),I)logP( \hat{y}=cls|\Delta I(S_{l}, N_l),I)
\end{equation}
where $L_{ID}$ is the Shannon entropy\cite{Shannon1948AMT} that uses the $\Delta I(S_{l}, N_l)$ for classification on image $I$, $I$ is sampled from the training set ,$\mathbb {C}$ is the number of the categories and $\hat{y}$ is the prediction based on the $\Delta I(S_{l}, N_l)$. By minimizing the $L_{ID}$, the pruned model can learn the important interaction eventually. Since $\Delta I(S_{l}, N_l)$ is non-negative, $L_{ID}$ can be effective guidance by avoiding the gradient explosion. Therefore, we integrate $\Delta I(S_{l}, N_l)$ with ground truth during the fine-tuning as shown in Figure \ref{fig:NPFI}.

\section{Experiments}
\begin{figure}[htbp]
\includegraphics[width=1.0\textwidth]{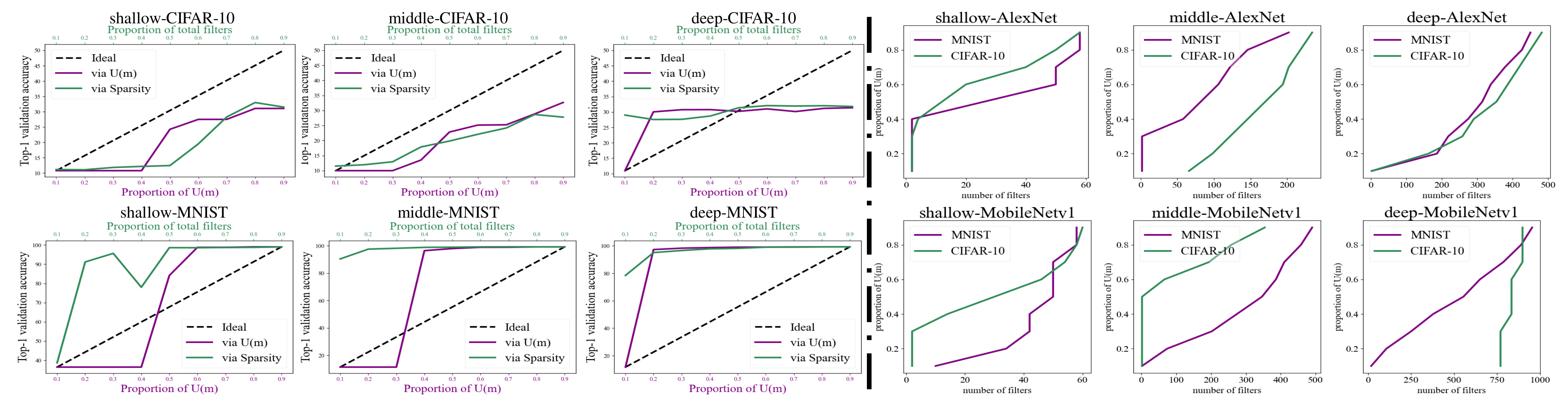}
\caption{Left: TOP-1 accuracy on CIFAR-10 (up) and MNIST (bottom) after pruning via filter utilization strength $U_l(m)$ and sparsity in AlexNet. Notably, we evaluate accuracy on the soft-pruned \cite{he2018soft} AlexNet without fine-tuning; Right: filter utilization strength in various layers of the AlexNet (up) and Mobilenetv1 (bottom) in CIFAR-10 (green) and MNIST (purple).}
\label{fig:um_val_um_diff_model_data}
\end{figure}
We empirically demonstrate the effectiveness of our proposed redundancy metric and the SNPFI on MNIST \cite{lecun1998gradient}, CIFAR-10 \cite{krizhevsky2009learning}, and ImageNet \cite{deng2009imagenet}. We implemented our method based on the publicly available Torch implementation for ResNet-50 \cite{he2016deep}, AlexNet \cite{krizhevsky2017imagenet}, and MobileNetv1 \cite{howard2017mobilenets}. As a commonly used lightweight network, MobileNetv1 is used to test the reliability of SNPFI when pruning on a sparse architecture. AlexNet and ResNet-50 are used to test the commonality of SNPFI with different densities of network connections.

\textbf{Baselines.} We mainly compare our approach with feature-based pruning: NS (Network slimming) \cite{liu2017learning} and NPPM \cite{gao2021network}. The NS is a one-shot pruning algorithm and requires no extract training after each pruning; the NPPM requires training for the feature selection network. To demonstrate the efficiency of SNPFI, weight-based pruning \cite{9041096,2021Neural,2019Towards,he2018soft} and Auto-ML methods \cite{2019arXiv190509717D,he2018amc} are included.

\textbf{Pruning.} For optimization of the SNPFI agent, we use the Adam algorithm to optimize the actor $\pi_\epsilon$ and the critic networks $\rho_\beta$. The learning rate is $10^{-4}$ for the actor and $10^{-5}$ for the critic. For fine-tuning, We utilize our proposed ID and wight it and the original loss as 0.5 and 1. For fine-tuning, we use SGD with a Nesterov momentum of 0.9. For AlexNet and MobileNetv1 on MNIST, 15 epochs are used with the initial learning rate of 5e-5; For MobieleNet-v1 and ResNet-50 on CIFAR10, 120 epochs are used with an initial learning rate of 8e-4; For MobieleNet-v1 and ResNet-50 on ImageNet, 160 epochs are used with an initial learning rate of 1e-3.

\subsection{Reliability of Filter-wise interaction}
%Filter utilization and interaction difference are the two principal applications of Filter interaction. The Filter utilization determines a suitable sparsity lower bound in constant time, while the interaction difference turns the pruned layer into a highly cohesive architecture.; the filters might interact actively (the red triangular region)
% Filters with higher importance are not always strongly related to the correct prediction, which means that the current redundancy criterion can not promise the pruned filters are subtle. In this situation, pruning the filter without considering the filter-wise interaction might ruin the performance. Therefore, filter-wise interaction is necessary when measuring redundancy. 

\textbf{Effectiveness of the filter utilization strength.} In the ideal, the number of filters has a linear relationship with the accuracy, which indicates that lower sparsity leads to better performance with promise. In reality, the accuracy doesn't monotonously increase with the proportion of total filters as shown in the shallow or middle layer in AlexNet (Figure \ref{fig:um_val_um_diff_model_data}-left). Thus, assigning all layers with the same sparsity is not effective. Since the accuracy monotonously increases with the filter utilization strength $U_l(m)$, a higher $U_l(m)$ can ensure better accuracy and is more effective than the sparsity. In this way, our reward function in Eq.(\ref{eq:reward}) can effectively alleviate the delayed reward problem. Meanwhile, the $U_l(m)$ converges around $0.5$ in diverse layers and applications as shown in Figure \ref{fig:um_val_um_diff_model_data}-left. Thus, we can estimate the sparsity lower bound $s_{lb}^l$ in Eq.(\ref{eq:slb}) with a uniform filter utilization strength $\theta$ for various layers. Since the filter-wise interactions are diverse among layers, the identical value of $U_l(m)$ implies different $s_{lb}^l$. We utilize this property to boost the optimization of the pruning plan as shown in Eq.(\ref{eq:opt_prune}). 

%\textbf{Effectiveness of the filter utilization strength.} 
% This part is about how to decide a U(m) for model by experience:As shown in Figure \ref{fig:um_val_um_diff_model_data}-left, filter utilization strength $U_l(m)$ in various layers are positively related to accuracy while negatively correlated to the validation loss, regardless of the depth. Hence, we can efficiently estimate the performance of the pruned via $U_l(m)$ without fine-tuning. Because all pruned models start to converge and become functional, the pruned model starts to thrive when $U_l(m)=0.6$. For layers with $U_l(m)\geqslant0.6$,  $m$ filters ensure the fundamental decision ability of the pruned model, and the rest $c_{out}^l-m$ filters are relatively subtle. Thus, we can roughly estimate the number of redundant filters by $U_l(m)$ and experience. 

\begin{figure}[htbp]
    
    \includegraphics[width=1.0\textwidth, height=0.3\textwidth]{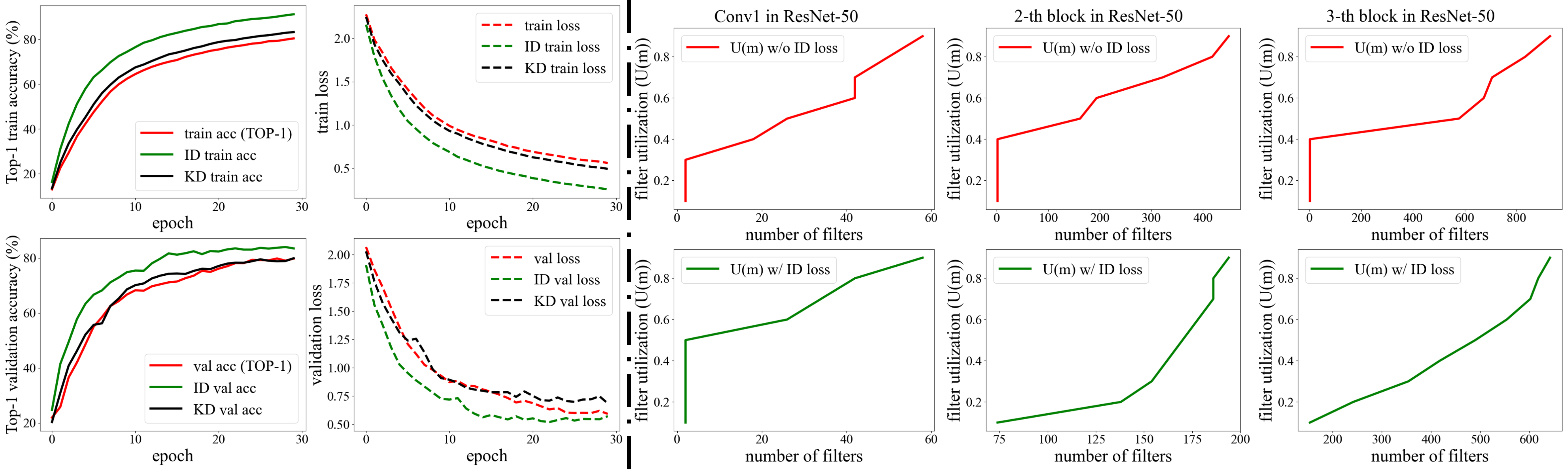}
    \caption{Left: The performance recovery outcomes by plain training \cite{ruder2016overview} (red), Knowledge distillation \cite{hinton2015distilling} (black), and our proposed interaction difference (green) when pruning around 60\% filters of MobileNetv1. Right: $U(m)$ of different layers in ResNet-50 (red) and the pruned ResNet-50 (green).}
    \label{fig:kd_id-umc}
\end{figure}

\textbf{Various interactive mode.} As shown in Figure \ref{fig:um_val_um_diff_model_data}-right, a model has three types of interactive modes: low channel first, multi-channel first, and high channel first. As the $U(m)$ of the shallow layer in AlexNet shown, the ‘low channel first’ is the interaction mode when layers’ $U(m)$ reach the inflection point with a few filters; As the $U(m)$ of AlexNet's the middle layer in AlexNet shown, the ‘multi-channel first’ is the interaction mode that layers’ $U(m)$ distinctly increase multiple times with the growing number of the filters; As the $U(m)$ of the deep layer in MobileNetv1 on the MNIST shown, the ‘high channel first’ is the interaction mode that layers’ $U(m)$ distinctly soar with a high number of channels. Since the interactive mode is diverse in the same model, scheduling aggressive pruning sparsity for a single layer can risk the generalization performance of the whole model. Specifically, even the identical model might behave variously when applied in different scenarios. As shown in Figure \ref{fig:um_val_um_diff_model_data}-right, MobileNetv1 tends to interact in ‘multi-channel first’ in MNIST but shifts from ‘multi-channel first’ to ‘low channel first’ and ends in ‘high channel first’ in CIFAR-10. Hence, our redundancy criterion is vital in real applications to prevent pruning the useful filters.

\textbf{Effectiveness of the interaction difference.} As shown in Figure \ref{fig:kd_id-umc}-left, our proposed interaction difference loss $L_{ID}$ is effective in weight recovery for aggressive pruning. Knowledge distillation \cite{hinton2015distilling} struggles to converge when the ground truth is intricate for training the pruned model. In contrast, the $L_{ID}$ steadily guides the pruned model to an increasing training accuracy without drastic fluctuation of validation loss. As shown in Figure \ref{fig:kd_id-umc}-right, the interaction difference can refine the interactive mode for the pruned model. Before pruning, the ResNet-50 tends to interact in ‘low channel first’ for shallower layers and in ‘multi-channel first’ for deeper layers. After fine-tuning with $L_{ID}$, all layers tend to interact in ‘multi-channel first’ mode, which means filters strongly cooperate and few redundancies exist.

\subsection{Pruning models on different datasets}\label{subsection:4.2}
\textbf{Pruning on CIFAR-10.} As shown in Table \ref{tab:RGBprune-cifar}, SNPFI reduces more than 60\% computation of the MobileNetv1 and ResNet-50 without a significant accuracy drop. In the deployment scenario, our method, nearly $2\times$ faster than the QPNN, can achieve comparable performance with QPNN on the FPGA device without any particular software accelerator \cite{yingge2020deep}. When compressing the highly computational network ResNet-50, SNPFI achieves the highest compression and accuracy among other state-of-the-art methods. Compared with AMC and TAS, SNPFI provides better network architecture by overcoming the delayed reward; When comparing with Greg-2, our method realizes 3.7\% more reduction and 2.54\% higher accuracy without iterative training to recover the weight; Without empowering the channel selection ability by manual modification of the network, our methods are more applicable and better performance than NS and NPPM.

\begin{table}[htbp]
    \small
    \caption{Pruning result on CIFAR10 (\ref{tab:RGBprune-cifar}) and ImageNet(\ref{tab:RGBprune-imgenet}). 'Acc' is the top-1 classification accuracy on the validation set,  and PR (Pruning Ratio) is computed by $1-\frac{Pruned\, FLOPs}{Original\, FLOPs}$.}
%     \begin{table}
%   \caption{Sample table title}
%   \label{sample-table}
%   \centering
%   \begin{tabular}{lll}
%     \toprule
%     \multicolumn{2}{c}{Part}                   \\
%     \cmidrule(r){1-2}
%     Name     & Description     & Size ($\mu$m) \\
%     \midrule
%     Dendrite & Input terminal  & $\sim$100     \\
%     Axon     & Output terminal & $\sim$10      \\
%     Soma     & Cell body       & up to $10^6$  \\
%     \bottomrule
%   \end{tabular}
% \end{table}

    \begin{subtable}[h]{0.45\textwidth}
        \caption{Pruning result on CIFAR10.}
        \label{tab:RGBprune-cifar}
        \centering
        \tabcolsep=0.5em
         \begin{tabular}{cccc}
         \toprule
         Method & Acc & Acc Drop & PR\\[0.5ex] 
         \midrule
         \multicolumn{4}{c}{MobileNetv1}\\
         \midrule
         QPNN (\citeyear{9041096})  & 91.53$\%$ & -0.22$\%$ & 37.50$\%$ \\ 
         NS (\citeyear{liu2017learning}) & 90.29$\%$ & 1.02$\%$ & 46.30$\%$ \\
         NPPM (\citeyear{gao2021network}) & 90.56$\%$ & 0.75$\%$ & 67.79$\%$ \\
         SNPFI(our) & \textbf{90.42}$\%$ & \textbf{0.89}$\%$ & \textbf{63.00}$\%$ \\[1ex]
         \midrule
         \multicolumn{4}{c}{ResNet-50}\\
         \midrule
         TAS \citeyear{2019arXiv190509717D} & 93.69$\%$ & -0.77$\%$ & 52.70$\%$ \\ 
         AMC (\citeyear{he2018amc}) & 91.90$\%$ & 1.83$\%$ & 50.00$\%$ \\
         Greg-2 (\citeyear{2021Neural}) & 93.36$\%$ & -0.44$\%$ & 60.80$\%$ \\
         NS (\citeyear{liu2017learning}) & 93.44$\%$ & -0.52$\%$ & 57.80$\%$ \\
         NPPM (\citeyear{gao2021network})& 93.48 $\%$ & -0.56$\%$ & 58.30$\%$ \\
         SNPFI(our) & \textbf{95.90}$\%$ & \textbf{-2.98}$\%$ & \textbf{64.50}$\%$ \\[1ex]
         \bottomrule
         \end{tabular}
    \end{subtable}
    \hfill
    \begin{subtable}[h]{0.45\textwidth}
        \caption{Pruning result on ImageNet.}
        \label{tab:RGBprune-imgenet}
        \centering
        \tabcolsep=0.5em
        \begin{tabular}{ cccc}
         \toprule
         Method & Acc &Acc Drop & PR\\ [0.5ex]
         \midrule
         \multicolumn{4}{c}{MobileNetv1}\\
         \midrule
         AMC (\citeyear{he2018amc}) & 73.00$\%$ & -2.94$\%$ & 56.00$\%$\\ 
         NS (\citeyear{liu2017learning})& 68.97$\%$ & 1.09$\%$&62.00$\%$\\
         SNPFI(our) & 71.93$\%$ & -1.87$\%$ & 52.50$\%$ \\[1ex]
         
         \midrule
         \multicolumn{4}{c}{ResNet-50}\\
         \midrule
         TAS (\citeyear{2019arXiv190509717D}) & 76.2$\%$ &  1.26$\%$ &43.50$\%$\\ 
         AMC (\citeyear{he2018amc}) & 76.11$\%$ & 1.35$\%$ & 50.00$\%$ \\
         Greg-2 (\citeyear{2021Neural}) & 76.13$\%$ & 1.32$\%$ &32.90$\%$\\
          LeGR (\citeyear{2019Towards})& 75.70$\%$& 1.76$\%$&42.00$\%$\\
         SFP (\citeyear{he2018soft}) & 74.61$\%$ & 2.85$\%$ & 41.80$\%$\\
         NPPM (\citeyear{gao2021network}) & 75.96$\%$ & 1.50$\%$ &56.00$\%$\\
         SNPFI(our) & \textbf{78.80}$\%$ & \textbf{-1.34}$\%$ &\textbf{53.60}$\%$\\[1ex]
         \bottomrule
         \end{tabular}
        
    \end{subtable}

\end{table}

\textbf{Pruning on ImageNet.} As shown in Table \ref{tab:RGBprune-imgenet}, SNPFI achieves the best accuracy when pruning generously on ResNet-50. When comparing with Auto-ML methods, SNPFI prunes a 10\% of Filters more than the TAS and performs a 2.39\% of accuracy higher than the AMC; When comparing with weight-based pruning methods, SNPFI prunes a 10\% of channels more than the Greg-2 and LeGR and performs exceeding 2\% of accuracy higher than those methods; When comparing with feature-based pruning methods, SNPFI out-performs the SFP and NPPM under comparable pruning ratio. When pruning on MobileNetv1, SNPFI realizes a comparable compression result with the AMC and NS.

\textbf{RGB model for single-band image.} According to Table \ref{tab:singleprune-MNIST} and \ref{tab:sparsityForRGB-GS}, we can experimentally prove that the architectural sparsity varies for the single-band and the multi-band image. On the one hand, different pruning methods can generalize well with less than 45\% of the origin when processing single-band images as shown in Table \ref{tab:singleprune-MNIST}, which is lower than the empirical value (60\% at most \cite{liu2017learning}).On the other hand, the redundancy does lie at the architecture level when processing single-band images by RGB model. As shown in Table \ref{tab:sparsityForRGB-GS}, applying SNPFI can decline 12\% higher of the computation overhead in the ResNet-50 for gray-scale image than the RGB image. The accuracy disparity between the RGB and the gray-scale input might be caused by the absence of colorful information in the gray-scale image, which is vital for object recognition.

\begin{table}[htbp]
\small
\caption{Pruning result on MNIST (\ref{tab:singleprune-MNIST}) and the ResNet-50's sparsity when processing the input image in the RGB and gray-scale modes (\ref{tab:sparsityForRGB-GS}).}
    \begin{subtable}[h]{0.5\textwidth}
        \centering
        \caption{Pruning result on MNIST.}
        \tabcolsep=0.5em
        \label{tab:singleprune-MNIST}
         \begin{tabular}{cccc}
             \toprule
             Method & Acc & Acc Drop & PR \\ [0.5ex] 
             \midrule
             \multicolumn{4}{c}{AlexNet}\\
              \midrule
             NS (\citeyear{liu2017learning}) & 99.00$\%$ & -0.04$\%$ & 51.20$\%$ \\
             NPPM (\citeyear{gao2021network}) & 99.13 & -0.17 & 55.10 \\
             SNPFI(our) & \textbf{99.14}$\%$ & \textbf{-0.18}$\%$ & \textbf{55.80}$\%$ \\[1ex]
              \midrule
             \multicolumn{4}{c}{MobileNetv1}\\
             \midrule
              NS (\citeyear{liu2017learning}) & 96.64$\%$ & 0.16$\%$ & 40.80$\%$ \\
             NPPM (\citeyear{gao2021network}) & 96.54$\%$ & 0.26$\%$ & 62.96$\%$ \\
             SNPFI(our) & \textbf{97.48}$\%$ & \textbf{-0.68}$\%$ & \textbf{59.10}$\%$ \\[1ex]
             \bottomrule
         \end{tabular}
    \end{subtable}
    \begin{subtable}[h]{0.5\textwidth}
        \centering
        \caption{The ResNet-50's sparsity when processing the RGB image and the gray-scale image in CIFAR10. We use the pruning ratio (PR) to represent the sparsity. The pruning is achieved by the SNPFT. We generate the gray-scale image by the red band information of the RGB image. The baseline model for processing gray-scale CIFAR10 is trained from the sketch with the same training setting for processing the original CIFAR10.  }
        \label{tab:sparsityForRGB-GS}
        \tabcolsep=0.5em
         \begin{tabular}{cccc}
             \toprule
             Image mode & ACC & Acc Drop & PR \\ [0.5ex] 
             \midrule
              RGB & 95.90$\%$ & -2.98$\%$ & 63.0$\%$ \\
             Gray-scale & 93.06$\%$ & -1.02\% & 75.51$\%$ \\
             \bottomrule
         \end{tabular}
    \end{subtable}
\end{table}

%\textbf{Pruning on MNIST.}As shown in table \ref{tab:singleprune-MNIST}, NPFI reduces nearly 60\% computation of the AlexNet and MobileNetv1 with few accuracy increase. When compression for better generalization, our method requires less computation but provides better accuracy than other feature-based pruning methods. What's more, all experimental methods achieved much higher pruning ratio than compression results for RGB images. \newline
\section{Conclusion}
In this paper, we discussed redundancy from the filter interaction aspect. On the one hand, our proposed redundancy metric can effectively predict the model's inherent sparsity and suggest the proper model for a real application. On the other hand, computing the filter utilization rate of our redundancy metric can be time-consuming. Since the number of filters grows with the depth of CNN, the combination number of the interaction can be enormous.  In the future, a considerable improvement in computation can be achieved by clustering on filters to evaluate the filter-wise utilization among distinct filters only.  

{\small
\bibliographystyle{plainnat}
\bibliography{main}
}

\end{document}